\documentclass[
]{ceurart}

\usepackage{listings}
\usepackage{tabularx}
\begin{document}

\copyrightyear{2025}
\copyrightclause{Copyright for this paper by its authors.
  Use permitted under Creative Commons License Attribution 4.0
  International (CC BY 4.0).}

\conference{Forum for Information Retrieval Evaluation, December 17-20, 2025, Varanasi, India}

\title{Adapting Small Language Models to Low-Resource Domains: A Case Study in Hindi Tourism QA}


\author[1]{Sandipan Majhi}[%
email=sandipan.majhi.24@kgpian.iitkgp.ac.in
]
\fnmark[1]
\cormark[1]
\author[2]{Paheli Bhattacharya}
\address[1]{Indian Institute of Technology Kharagpur, India}
\address[2]{Bosch Research and Technology Centre, Bangalore, India}

\cortext[1]{Corresponding author.}
\fntext[1]{Work done during the internship at Bosch Research and Technology Centre, Bangalore, India}

\begin{abstract}
Domain-specific question answering in low-resource languages faces two key challenges: scarcity of annotated datasets and limited domain knowledge in general-purpose language models. In this work, we present a multi-stage finetuning strategy to adapt lightweight language models to the Hindi tourism domain by leveraging both original and synthetic training data. Synthetic question–answer pairs are generated using large LLMs (LLaMA-70B, Phi-14B) and used to augment the limited original dataset. We explore several training methodologies and analyze their impact on domain generalization. Our results demonstrate that large models can efficiently generate synthetic data, while small models can effectively adapt to it, offering a scalable pathway for low-resource, domain-specific QA.
\end{abstract}


\maketitle

\section{Introduction}

Large language models (LLMs) have significantly advanced natural language generation, understanding, and reasoning. Despite their success, adapting these models to domain-specific applications remains challenging due to two main factors: (i) general-purpose LLMs often lack specialized domain knowledge, and (ii) high-quality annotated datasets are scarce and expensive to obtain. The cost and time demands of manual annotation have therefore driven interest in synthetic data as a scalable alternative.

LLMs, owing to their broad pre-training, can act as effective knowledge bases \cite{sun-etal-2024-head, wang2021can} and have been shown to produce high-quality synthetic question–answer (QA) pairs \cite{scaria-etal-2024-good, yuan-etal-2023-selecting, schmidt-etal-2024-prompting}. Synthetic datasets have further demonstrated utility in addressing the limitations of low-resource domains \cite{tengler2025exploring, hakam2024human}, enabling the creation of specialized training resources that would otherwise be infeasible. This has opened a practical avenue for domain adaptation, especially in fields where curated open-source datasets are extremely limited.

At the same time, the emergence of lightweight language models provides new opportunities for efficient domain adaptation. Smaller models are cheaper to finetune, faster at inference, and easier to deploy in resource-constrained environments. While very large LMs are well-suited for generating synthetic training data, compact models are more practical for downstream deployment. Thus, combining synthetic data generation from large models with targeted finetuning of smaller ones represents a promising strategy for building effective, domain-specific QA systems.

In this work, we investigate this paradigm in the context of Hindi tourism, a domain where both language resources and annotated datasets are limited. We generate synthetic QA pairs using large LMs (LLaMA-70B and Phi-14B) and finetune a smaller model (LLaMA-8B) to evaluate its performance. Beyond simple finetuning, we explore mixed-training methodologies that combine synthetic and general-domain data, analyzing their effect on robustness and domain generalization. Our contributions are threefold: (i) a data augmentation strategy tailored to low-resource, domain-specific QA, (ii) empirical evidence that synthetic data can effectively adapt lightweight models, and (iii) a comparative analysis of training strategies to identify setups that balance efficiency with domain performance.
\section{Related Work}
\vspace{-2mm}

Generating \textbf{high-quality synthetic question answers} using large language models (LLMs) has been a key focus of recent research \cite{scaria-etal-2024-good, yuan-etal-2023-selecting, schmidt-etal-2024-prompting}. Studies done by \citet{chia-etal-2022-relationprompt} and \citet{liu-etal-2024-unleashing-power} have shown that zero-shot prompting can be highly effective for creating high-quality, structured synthetic data. However, generating synthetic question-answer pairs can sometimes result in unintended redundancy. Studies such as \citet{yadav-etal-2024-explicit} suggest that exploring different sampling techniques could introduce greater diversity, which may be beneficial for downstream tasks. The primary goal of generating synthetic question-answer pairs is to improve model performance on question-answering tasks. Prior work by \citet{chowdhury-chadha-2024-generative} demonstrated how synthetic data, particularly from "in-the-wild" sources, can lead to performance gains and help achieve natural distribution shifts. Similarly, \citet{kramchaninova-defauw-2022-synthetic} validated the effectiveness of combining synthetic data with original training data, showing that this approach consistently outperforms models trained exclusively on non-synthetic data, especially on domain-specific test sets. Another study by \citet{harsha-etal-2025-synthetic} on the use of synthetic data within the financial domain confirmed its effectiveness in boosting question-answering performance in specialized fields.

To achieve performance improvements on downstream question-answering tasks, several studies have investigated \textbf{different methods for finetuning models} using a combination of synthetic and original training datasets.  \citet{namboori2023gemquad} proposed a finetuning approach that involves first training on the synthetic data and then on the original training set, arguing that a model should perform better if it is well-conditioned to a high-quality dataset. Conversely, \citet{chada-natarajan-2021-fewshotqa} showed performance improvements by finetuning first on the original training data and then on a small, additional amount of synthetic data. Other studies, including \cite{chen-etal-2024-minprompt, ushio-etal-2023-empirical, schmidt-etal-2024-prompting}, also demonstrate that continued finetuning on a small amount of synthetic data can lead to a significant performance uplift in question answering. A study by \citet{gurgurov-etal-2024-adapting} utilized synthetic data curation by translating English data into other low-resource languages and performing continued pretraining, illustrating how synthetic data can aid in model alignment for new domains.

Researchers have already created several benchmarks based on \textbf{synthetic datasets, particularly for low-resource domains}. The Indic-QA Benchmark \cite{singh-etal-2025-indic} used synthetic data generation techniques to create question-answering datasets in 11 Indian languages. Similarly, IndiSentiment140 \cite{kumar-etal-2024-indisentiment140} is another such dataset that uses machine translation to generate sentiment analysis datasets across 22 Indian languages. The IndicXTREME Benchmark \cite{doddapaneni-etal-2023-towards} also leveraged machine translation to create new synthetic datasets from existing English data.

\section{Methodology}
\label{sec:method}
This study investigates the impact of synthetic data augmentation on the performance of small language models for long-form question-answering task. 

\noindent
\textbf{Synthetic Data Generation}: As shown in Table~\ref{tab:dataset_stats},  we utilize contexts from the training set to generate additional question-answer pairs. To achieve this, we employ larger LLMs to create a new corpus of question answer pairs using training set contexts using few shot prompts. 

\begin{figure}[t]
  \centering
  \fbox{\includegraphics[width=0.8\linewidth]{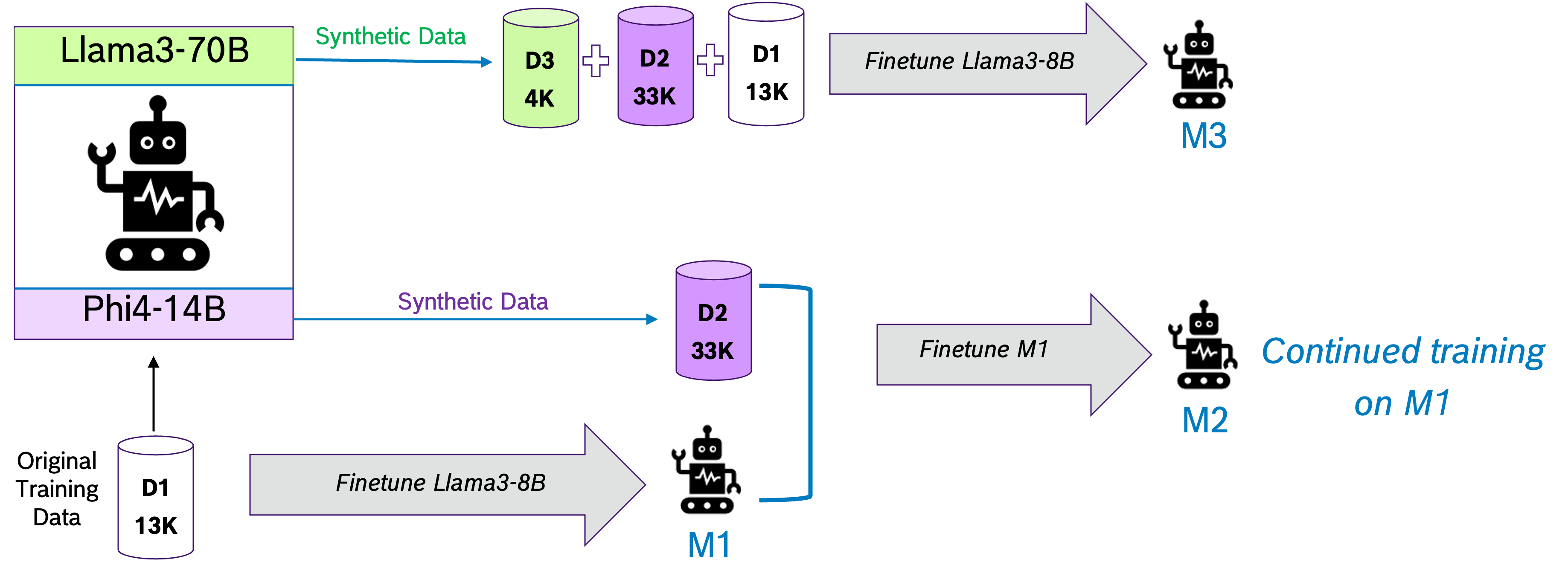}}
  \caption{An overview of the two-phased experimental procedure, including synthetic data generation, followed by mixed fine-tuning of a smaller language model on the augmented Hindi-language dataset.}
  \label{fig:workflow_image}
\end{figure}

\noindent
\textbf{Finetuning SLMs}: We train SLMs using the synthetic data and available training data. A comprehensive representation of the workflow has been presented in Figure ~\ref{fig:workflow_image}. We follow a mixed-training strategy:

\noindent
$\bullet$ Baseline Model: We first finetune a small language model on the already available data, to get the baseline finetuned model. 

\noindent
$\bullet$ Continued Finetuning: We use the synthetic data and continue finetuning the baseline finetuned model, to produce a continued finetuned model.

\noindent
$\bullet$ Multi-Source Finetuning: In this setting, we use all of the available training data and the augmented synthetic data to finetune the SLM.


\section{Dataset}
\label{sec:dataset}
This study utilizes the Varanasi Tourism in Question Answer System (VATIKA) dataset published by \citet{gatla2025tourism}, a publicly available resource in Forum for Information Retrieval Evaluation (FIRE) 2025. This Hindi-language dataset consists of instances where each context is paired with one or more related question-answer pairs. The answers were typically long-form and abstractive in nature, utilizing the provided context as their source of information. The originally published dataset had three splits, namely, train, validation and Test Data-1. There is a held-out test set which was provided as a part of the shared task. It only had contexts and questions and not the gold standard answers. We refer this dataset as Test Data-2.  To provide a comprehensive overview of the dataset we present its key statistics in Table ~\ref{tab:dataset_stats}. We see that on average the question and answer lengths in the VATIKA dataset is 13 and 16 words respectively and there are about 2-3 question-answer pairs per contexts. The synthetic data on average produces approximately 3 more QA pairs on average and has similar question and answer length distribution.
\begin{table}[t]
    \centering
    \scriptsize
    \resizebox{\textwidth}{!}{
        \begin{tabular}{lrrrrr}
        \toprule
        \textbf{Split} & \textbf{Contexts} & \textbf{QA Pairs} & \textbf{QA/Context} & \textbf{Ques. length} & \textbf{Ans. length} \\
        \midrule
        Train & 5244 & 13092 & 2.50 & 12.64 & 16.10 \\
        Validation & 1134 & 2798 & 2.47 & 12.57 & 16.07 \\
        Test Data-1 & 1143 & 2902 & 2.53 & 12.68 & 16.16 \\
        Test Data-2 & 430 & 1196 & 2.78 & 16.40 & -- \\
        \textbf{Synthetic Data} & 5244 & 37259 & 5.11 & 12.16 & 18.63 \\ 
        \bottomrule
        \end{tabular}
        }
    \caption{Key statistics of the VATIKA dataset, including question-answer pairs, average sentence lengths, and the number of contexts in each split. We could not provide the answer length of Test Data-2 as it was held-out.}
    \label{tab:dataset_stats}
\end{table}

\section{Experimental Settings}
\label{sec:expr}
\textbf{Synthetic Data Generation}: For generating synthetic data, LLAMA-3.1-70B\cite{dubey2024llama} and Phi-4-14B\cite{abdin2024phi} were utilized in few shot prompt format. The new corpus of question-answer pairs contained around 4,000 instances generated by LLAMA-3.1-70B and 33,000 instances generated by Phi-4-14B. We use $temperature = 0.7$ and $top-p = 0.9$ for synthetic data generation for both the models.

\noindent
\textbf{Model Finetuning}: We finetune LLAMA-3.1-8B\cite{dubey2024llama} using the strategies described in Section~\ref{sec:method}. We experiment with three distinct model configurations as follows. The hyperparameters are in Table ~\ref{tab:finetuning-parameters}.

\noindent
$\bullet$ M1: Baseline Model: LLAMA-3.1-8B was exclusively fine-tuned for 4 epochs on the original 13,092 training instances.

\noindent
$\bullet$ M2: Continued Fine-Tuning: The 2 epoch trained baseline model (M1) underwent a second phase of fine-tuning for another 2 epochs on a 33,000-instance synthetic dataset generated by Phi-4-14B.

\noindent
$\bullet$ M3: Multi-Source Fine-Tuning: This model was fine-tuned for 4 epochs on a combined 50,000-instance dataset, which included the original 13,000 training instances along with synthetic data from two distinct large language models: 33,000 instances from Phi-4-14B and 4,000 instances from LLAMA-3.1-70B.


\begin{table}[!thb]
    \centering
    \begin{tabular}{lc}
        \toprule
        \textbf{Parameter} & \textbf{Value} \\
        \midrule
        Max sequence length & 4096 \\
        Per-device train batch size & 2 \\
        Gradient accumulation steps & 4 \\
        Warmup steps & 5 \\
        Learning rate scheduler type & "cosine" \\
        Number of epochs & 4 \\
        \bottomrule
    \end{tabular}
    \caption{Fine-Tuning Parameters for LLAMA-3.1-8B.}
    \label{tab:finetuning-parameters}
\end{table}

\noindent
\textbf{Evaluation}: We report our model's performance using token-based metrics, ROUGE-L\footnote{\url{https://huggingface.co/spaces/evaluate-metric/rouge}} and BLEU\footnote{\url{https://huggingface.co/spaces/evaluate-metric/bleu}}. In our implementation of the ROUGE-L scores presented in Table~\ref{tab:results} we modify  of the default tokenization function to incorporate Hindi words and characters. For the semantics-based metric, BERTScore \footnote{\url{https://huggingface.co/spaces/evaluate-metric/bertscore}} over predicted answers and gold answers on validation and test splits.


\section{Results and Analysis}
In this section, we first provide our evaluation of the different training strategies on Validation and Test Data-1. Then, we present the organizers' evaluation on Test Data-2.

\noindent
\textbf{Validation and Test Data-1}: Our experiments presented in Table~\ref{tab:results}, revealed several key findings regarding model training and distribution robustness. First, as shown in Table~\ref{tab:results}, the model trained only on the original data (M1[\ref{sec:expr}]) performed best on the development set, while the model with combined original and synthetic data (M3[\ref{sec:expr}]) excelled on Test Data-1. This disparity highlights the models' sensitivity to distribution shifts.

However, the combined multi-source model (M3[\ref{sec:expr}]) underperformed in BLEU-2, which is a precision based metric. A potential reason is that combining multi-source data may introduce conflicting answers for similar questions, leading to ambiguity and performance degradation.

\noindent
\textbf{Held-out Test Data-2}: Table~\ref{tab:fire_results} demonstrates our method's performance on the proprietary and undisclosed Test Data-2 split. Our model configurations has consistent top rankings with M2[\ref{sec:expr}] outperforming other models in QA-F1, a second-place ranking in BLEU-2, a third-place ranking in BLEU-1 and fourth-place ranking in ROUGE-1 and ROUGE-2. The results indicate that supplementing models with a large quantity of high-quality synthetic data can not only improve performance on downstream tasks but also significantly enhance their robustness to unseen data.

A crucial insight comes from the two-stage trained model (M2[\ref{sec:expr}]). Despite its moderate performance on Test Data-1, it achieved superior BLEU-2 and QA-F1 scores on the held-out Test Data-2 (Table~\ref{tab:fire_results}). This suggests that late exposure to synthetic data is effective for building distribution robustness.

\noindent
\textbf{Analysing the Synthetic data}: Table~\ref{tab:dataset_stats} outlines the quantitative differences between the original and synthetic datasets, while Table~\ref{tab:gen_comparison} presents a qualitative comparison of the question-answer (QA) pairs they generated. The outputs from LLAMA-3.1-70B and Phi-4-14B demonstrate considerable overlap, underscoring the critical importance of a data selection stage to filter for quality. A potential direction for future research in developing such quality checks is the joint evaluation of both question and answer within each synthetic pair.

\begin{table}[t]
    \centering
    \scriptsize
    \resizebox{\textwidth}{!}{
    \begin{tabular}{llcccccc}
        \toprule
        \textbf{Model} & \textbf{Settings} & \multicolumn{3}{c}{\textbf{Validation Data}} & \multicolumn{3}{c}{\textbf{Test Data-1}} \\
        \cmidrule(lr){3-5} \cmidrule(lr){6-8}
        & & \textbf{Rouge-L} & \textbf{BLEU} & \textbf{BERTScore} & \textbf{Rouge-L} & \textbf{BLEU} & \textbf{BERTScore} \\
        \midrule
        M1 & Orig 13k & \textbf{0.897} & \textbf{0.799} & \textbf{0.971} & 0.917 & 0.838 & 0.976 \\
        M2 & M1 + 33k & 0.893 & 0.790 & 0.969 & 0.911 & 0.828 & 0.975 \\
        M3 & All 50k & 0.892 & 0.791 & 0.968 & \textbf{0.922} & \textbf{0.849} & \textbf{0.978} \\
        \bottomrule
    \end{tabular}
    }
    \caption{Performance of our model settings on Dev set and Test Data-1 splits of publicly available VATIKA Dataset.}
    \label{tab:results}
\end{table}

\begin{table}[t]
    \centering
    \resizebox{\textwidth}{!}{%
    \begin{tabular}{@{}l c l c l c l c l c l c@{}}
        \toprule
        \multicolumn{2}{c}{\textbf{BLEU-1}} & 
        \multicolumn{2}{c}{\textbf{BLEU-2}} & 
        \multicolumn{2}{c}{\textbf{ROUGE-1}} & 
        \multicolumn{2}{c}{\textbf{ROUGE-2}} & 
        \multicolumn{2}{c}{\textbf{ROUGE-L}} & 
        \multicolumn{2}{c}{\textbf{QA-F1}} \\
        \cmidrule(lr){1-2} \cmidrule(lr){3-4} \cmidrule(lr){5-6} \cmidrule(lr){7-8} \cmidrule(lr){9-10} \cmidrule(lr){11-12}
        \textbf{Teams} & \textbf{Score} &
        \textbf{Teams} & \textbf{Score} &
        \textbf{Teams} & \textbf{Score} &
        \textbf{Teams} & \textbf{Score} &
        \textbf{Teams} & \textbf{Score} &
        \textbf{Teams} & \textbf{Score} \\
        \midrule
        Scalar1 & 63.7 & Scalar1 & 41.2 & IRel3 & 0.0824 & IRel3 & 0.0467 & IRel3 & 0.0824 & \textbf{OUR\_M2} & \textbf{0.576} \\
        IRel3 & 61.5 & \textbf{OUR\_M2} & \textbf{38.7} & AiNauts1 & 0.0818 & AiNauts1 & 0.0454 & AiNauts1 & 0.0818 & \textbf{OUR\_M3} & \textbf{0.566} \\
        \textbf{OUR\_M3} & \textbf{60.9} & \textbf{OUR\_M1} & \textbf{38.4} & CSE\_SVNIT3 & 0.0790 & CSE\_SVNIT3 & 0.0452 & CSE\_SVNIT3 & 0.0790 & \textbf{OUR\_M1} & \textbf{0.562} \\
        \textbf{OUR\_M2} & \textbf{60.1} & IRel3 & 36.4 & \textbf{OUR\_M2} & \textbf{0.0777} & \textbf{OUR\_M1} & \textbf{0.0424} & \textbf{OUR\_M1} & \textbf{0.0777} & IRel3 & 0.551 \\
        NamasteNLP1 & 59.9 & AiNauts2 & 33.2 & \textbf{OUR\_M1} & \textbf{0.0773} & \textbf{OUR\_M2} & \textbf{0.0415} & \textbf{OUR\_M3} & \textbf{0.0773} & Scalar1 & 0.505 \\
        AiNauts1 & 56.5 & NamasteNLP1 & 31.7 & \textbf{OUR\_M3} & \textbf{0.0763} & MUCS1 & 0.0438 & \textbf{OUR\_M2} & \textbf{0.0763} & IRel2 & 0.461 \\
        AiNauts2 & 55.8 & IRel1 & 30.5 & MUCS1 & 0.0759 & \textbf{OUR\_M3} & \textbf{0.0411} & MUCS1 & 0.0759 & AiNauts1 & 0.453 \\
        Scalar-2 & 52.3 & IRel2 & 28.6 & MUCS3 & 0.0685 & MUCS3 & 0.0411 & MUCS3 & 0.0685 & CSE\_SVNIT1 & 0.433 \\
        IRel1 & 51.2 & AiNauts1 & 27.2 & NamasteNLP3 & 0.0685 & NamasteNLP3 & 0.0392 & NamasteNLP3 & 0.0685 & IRel1 & 0.417 \\
        \textbf{OUR\_M1} & \textbf{49.5} & \textbf{OUR\_M3} & \textbf{27.0} & NamasteNLP2 & 0.0626 & NamasteNLP2 & 0.0360 & NamasteNLP2 & 0.0626 & CSE\_SVNIT2 & 0.394 \\
        \bottomrule
    \end{tabular}
    }
    \caption{Official results from the FIRE VATIKA Competition were tested on a private Test Data-2 split. The different model configurations for our method (VA-BO-INTERN) have been highlighted in bold.}
    \label{tab:fire_results}
\end{table}

\begin{table}[t]
\centering
\scriptsize
\resizebox{\textwidth}{!}{%
\begin{tabular}{@{}p{2.5cm} p{12cm}@{}}
\toprule
\textbf{Model} & \textbf{Questions and Answers (translated from Hindi for representation)} \\
\midrule
\textbf{Context} & 
Bhagirath Kund is 14.1 kilometers away from Pt. Deen Dayal Upadhyaya Railway Station. 
It can be reached by taxi, cab, or bus. 
This station was previously known as Mughalsarai and is one of India’s major railway junctions. 
The journey offers views of Varanasi’s historic lanes and ghats and provides an experience of its cultural heritage. \\
\midrule
\textbf{Original Training Data} & 
Q1: How far is Bhagirath Kund from Pt. Deen Dayal Upadhyaya Railway Station? \newline 
A1: Bhagirath Kund is 14.1 kilometers away from the station. \newline\newline
Q2: How can one reach Bhagirath Kund from Pt. Deen Dayal Upadhyaya Railway Station? \newline 
A2: One can reach Bhagirath Kund by taxi, cab, or bus services from the station. \\
\midrule
\textbf{LLAMA-3.1-70B} & 
Q1: What was the previous name of Pt. Deen Dayal Upadhyaya Railway Station? \newline 
A2: It was earlier known as Mughalsarai. \newline\newline
Q1: Is Pt. Deen Dayal Upadhyaya Railway Station an important railway junction? \newline 
A2: Yes, it is one of India’s major railway junctions. \newline\newline
Q3: What kind of views are seen on the journey to Bhagirath Kund? \newline 
A3: The journey offers views of Varanasi’s historic lanes and ghats. \newline\newline
Q4: What kind of experience do devotees get on this journey? \newline 
A4: Devotees experience Varanasi’s cultural heritage, enhancing the importance of the pilgrimage. \\
\midrule
\textbf{Phi-4-14B} & 
Q1: What was the old name of Pt. Deen Dayal Upadhyaya Railway Station? \newline 
A1: Its old name was Mughalsarai. \newline\newline
Q2: In which category does Pt. Deen Dayal Upadhyaya Railway Station fall? \newline 
A2: It is one of India’s major railway junctions. \newline\newline
Q3: What kind of views does the journey to Bhagirath Kund provide? \newline 
A3: It provides views of Varanasi’s historic lanes and ghats. \newline\newline
Q4: What kind of experience do devotees encounter on the journey? \newline 
A4: Devotees experience Varanasi’s cultural heritage, enhancing the significance of the site. \newline\newline
Q5: Where is Pt. Deen Dayal Upadhyaya Railway Station located? \newline 
A5: It is located in Varanasi. \\
\bottomrule
\end{tabular}}
\caption{Examples of synthetic question–answer pairs from LLAMA-3.1-70B and Phi-4-14B, along with training data. 
The original dataset is in Hindi; the questions and answers are translated into English for representation.}
\label{tab:gen_comparison}
\end{table}

\section{Conclusion and Future Work}

In this work, we proposed a multi-stage finetuning strategy for lightweight language models in the Hindi tourism domain, leveraging both original and synthetic training data. Models trained with continued finetuning-first on original data, then on synthetic data-consistently outperformed alternative approaches. This staged exposure allows the model to retain grounding in authentic data while benefiting from the scale and diversity of synthetic examples, improving robustness in domain-specific question answering. We also found that indiscriminate or excessive mixing of multi-source synthetic data can degrade performance, highlighting the importance of careful curation and controlled integration in low-resource settings.

As a future work our approach can be extended to other low-resource languages to test its generalizability. Future work also includes systematic evaluation of synthetic data quality, potentially using LLM-based filtering methods. Overall, this study demonstrates that large models can generate synthetic data, but small models can effectively adapt to it, enabling scalable and robust QA systems in low-resource domains.







\section*{Declaration on Generative AI}
 During the preparation of this work, the author(s) used Gemini-Flash 2.5 in order to rectify: Grammar, spelling check and evaluate microstructure. After using these tool(s)/service(s), the author(s) reviewed and edited the content as needed and take(s) full responsibility for the publication’s content. 

\bibliography{references}




\end{document}